\documentclass[a4paper, 10 pt, conference]{hsmr} 
\usepackage[utf8]{inputenc}
\IEEEoverridecommandlockouts                       
\usepackage{mathtools}

\usepackage[utf8]{inputenc}
\usepackage{mathtools}
\usepackage{geometry}
\usepackage{comment}
\usepackage[labelsep=space]{caption}
\usepackage{newtxtext,newtxmath}   
\usepackage{authblk}
\usepackage{pgfplots}
\usepackage{graphicx}
\usepackage{gensymb}
\usepackage{array}
\usepackage{stfloats}
\usepackage{soul}
\usepackage{cite}                  
\usepackage[colorlinks,urlcolor=blue]{hyperref}   
\usepackage{cleveref}              
\pgfplotsset{compat=newest} 
 
\title{\LARGE \bf
Soft Eversion Robots for Colonoscopy: Challenges, Open Problems, and Emerging Solutions} 
\author{\Large Cem Suulker}
\author{\Large Thomas Mack} 
\author{\Large Qingzheng Cong}
\author{\Large Neri Niccolò Dei} 
\author{\\\vspace{-1pt}\Large Reza Kashef}
\author{\Large Mohammad Sheikh Sofla}
\author{\Large Kaspar Althoefer} 
\affil{\normalsize\textit{School of Engineering and Materials Science, Queen Mary University of London, United Kingdom.}\\ \normalsize\textit{c.suulker@qmul.ac.uk}\vspace{-0.04\linewidth}}

\begin{document}
\bstctlcite{IEEEexample:BSTcontrol}
\maketitle
\thispagestyle{empty}
\pagestyle{empty}

\section*{INTRODUCTION}



Conventional colonoscopy remains limited by patient discomfort and procedural risks, motivating research into compliant robotic alternatives \cite{white2017cancer}. Eversion robots, which advance via pressure-driven tip growth, eliminate sliding friction against the colon wall and offer less intrusive traversal approach. However, no existing design simultaneously satisfies all clinical requirements. This paper examines four colonoscopy application eversion robot architectures, identifies the key trade-offs each reveals, and provides design guidance for future studies.

 Table I summarises the key anatomical constraints derived from existing datasets \cite{alqarni2024experimental,suulker2026state} against the performance of four recent eversion robot designs \cite{kim2025soft,davy2024vine,shi2025design,giri2025inchigrab}. Specifically colon length of 1.0--2.1 m, minimum luminal diameter of 26 mm in the sigmoid colon, bending angles up to 70° across 3–7 sharp bends, and a 2.3--3.8 mm working channel requirement \cite{ciuti2020frontiers}.


\begin{figure*}
    \centering
    \includegraphics[width=\linewidth]{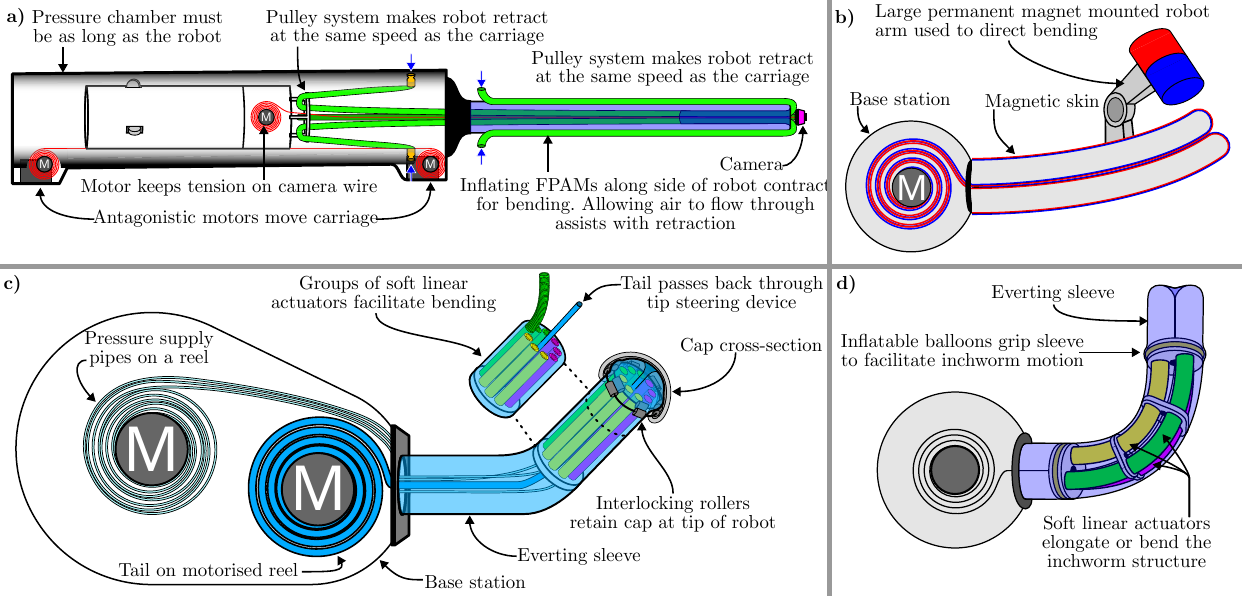}
    \caption{Eversion robots explored in this paper: a) Kim et al. \cite{kim2025soft} b) Davy et al. \cite{davy2024vine} c) Shi et al. \cite{shi2025design} d) Giri et al. \cite{giri2025inchigrab}}
    \label{fig:fig}
    \vspace{-0.04\linewidth}
\end{figure*}

\begin{table*}[!b]
    \centering
    \caption{Benchmarking of State-of-the-Art Eversion Robots against Clinical Colonoscopy Requirements}
    \label{tab:benchmarking_works}
    \begin{tabular}{|m{2.5cm}|m{2.8cm}|c|c|c|c|}
        \hline
        \textbf{Requirement} & \textbf{Anatomical Target} 
        & \textbf{Kim et al. \cite{kim2025soft}} 
        & \textbf{Davy et al. \cite{davy2024vine}} 
        & \textbf{Shi et al. \cite{shi2025design}} 
        & \textbf{Giri et al. \cite{giri2025inchigrab}} \\
        \hline \hline
        
        \textbf{Length} 
        & 1.5 -- 2.1 m \cite{alqarni2024experimental} 
        & 0.9 m $\times$ 
        & 0.25 m $\times$ 
        & 1.6 m \checkmark 
        & 1.45 m \\ 
        \hline
        
        \textbf{Diameter} 
        & $\leq$ 26 $\pm$ 4 mm \cite{suulker2026state} 
        & 25 mm \checkmark 
        & 8 mm \checkmark 
        & 18 mm \checkmark 
        & 27 mm \checkmark \\ 
        \hline
        
        \textbf{Maximum Bending} 
        & $\geq$ 52$^\circ$ (SD 22.4\degree{}) \cite{suulker2026state} 
        & $\sim85^\circ$(equiv.)\checkmark
        & N/A 
        & 201.8$^\circ$ \checkmark 
        & 130$^\circ$ \checkmark \\ 
        \hline
        
        \textbf{Payload} 
        & Camera / Tools \cite{suulker2026state} 
        & Cap-free \checkmark 
        & None $\times$ 
        & Rigid Cap $\times$ 
        & None $\times$ \\ 
        \hline
        
        \textbf{Material} 
        & Flexible / Soft 
        & Ripstop Nylon 
        & LDPE 
        & PU-Fabric 
        & Stretchlon \\ 
        \hline
    \end{tabular}
    
    \vspace{2mm}
    \small \textit{Note: \checkmark indicates the requirement is met for the majority of patients; $\times$ indicates a limitation in clinical applicability.}
\end{table*}

\section*{CURRENT EVERSION ROBOT SOLUTIONS AND THEIR LIMITATIONS}

Fig. \ref{fig:fig} highlights recent eversion robot studies for colonoscopy, while Table \ref{tab:benchmarking_works} shows that, despite strong overall performance, key limitations remain, particularly in payload delivery.

Fig. \ref{fig:fig}a) Eversion robot with multifunctional artificial muscles (green) that can contract upon inflation for bending or to assist with retraction. The base station must be at least the double of maximum length of the robot due to how it retracts, making it at least 2.1 m to reach at around 0.9 m into the colon. The small camera sits at the tip of the robot without making contact with the colon wall.

Fig. \ref{fig:fig}b) A magnetically polarised everting sleeve directed by an external magnet mounted on a robot arm. 
This eliminates onboard steering complexity but requires 
precise external magnet positioning, and the robot carries no payload.

Fig. \ref{fig:fig}c) A tip-steering device is retained at the tip by interlocking rollers with a rigid cap. The tip steering is performed by soft pneumatic linear actuators. While the tip-steering mechanism allows for bending angles that exceed the requirements for colon navigation, the rigid cap undermines the softness and the feasibility.

Fig. \ref{fig:fig}d) An inchworm  mechanism inside an eversion robot for bending and repositioning without external rigid fixtures. This approach does not allow for payload delivery.
The inchworm itself remains rigid, limiting the diameter of constriction that the robot can pass through.

\section*{DISCUSSION AND FUTURE DESIGN GUIDANCE}
\textbf{Material:}
For colonoscopy applications, the body of the eversion robot must be biocompatible, as it comes into direct contact with the colon wall. The material should also be highly flexible to enable navigation with minimal interaction forces (i.e., low bending stiffness), and low activation pressure so that, in the event of leakage or failure, pressurized air does not risk tissue damage.

In this context, recent literature highlights TPU and fabric-based materials as particularly promising for eversion robots, owing to their flexibility and low eversion pressure \cite{suulker2026state}, but quantitative benchmarking of safe eversion pressures against tissue contact damage thresholds remains an open problem for clinical development.

\textit{Key takeaway:} Thin materials, such as TPU or low thickness coated fabrics, should be selected for the robot sleeve to enable low bending stiffness, reduced actuation pressure, and improved navigation performance; both also offer biocompatible options.



\textbf{Steering Strategy:}
Pre-shaped robots are clinically impractical as they cannot adapt to real-time deformations or varied anatomies. Similarly, rigid steering methods \cite{shi2025design} compromise the robot’s compliance, a critical advantage for navigating variable luminal diameters. 

A more suitable strategy is tip-based soft steering, where actuation is localized at the robot’s tip, preserving compliance while enabling traversal of sharp, high-curvature bends commonly found in the colon. Enhancing the robot’s environmental guided growth and navigation capability by deliberately creating easy-buckling points offers another promising approach \cite{suulker2026enabling}.


Steering performance should be evaluated using appropriate metrics and realistic phantoms. Bending angle alone is insufficient and should be complemented by bending radius. Rigid phantoms are not representative, as they unrealistically transfer reaction forces and simplify environment-guided traversal. In contrast, the colon deforms under load, forming pockets that can trap the robot before sufficient reaction forces develop \cite{suulker2026state,suulker2026enabling}. Porcine colons are also suboptimal, as they are generally easier to traverse than the human colon.

\textit{Key takeaway:} Tip-based steering methods \cite{shi2025design} are effective but must preserve compliance. Passive strategies that enhance environment-guided traversal \cite{suulker2026enabling} show strong potential, and validation should be performed using flexible colon phantoms.

\textbf{Payload Delivery:}
Table \ref{tab:benchmarking_works} exposes a trilemma: payload delivery, compact base system, and full sleeve compliance cannot currently be achieved simultaneously. Existing approaches for delivering tools that use large rigid caps are suboptimal, as they increase tip friction against the colon wall, reduce flexibility, and pose a risk of tissue damage, thus defeating the core advantages of eversion robots.

Soft cap-based solutions \cite{suulker2023softcap,suulker2024deformable} offer an improvement by preserving the robot’s compliance, but they still generate unwanted friction. The ideal payload-carrying method should explore the internal channel as a route to keeping payloads at the tip of the eversion robot \cite{kim2025soft}. These methods preserve the robot's frictionless behavior and offer a better solution for clinical applications.

\textit{Key takeaway:} Payload delivery through the robot’s internal channel is effective \cite{kim2025soft}, but solutions that avoid bulky base systems are required.

\textbf{Conclusion and Future Work:}
Across the reviewed designs, the main challenge for colonoscopy eversion robots is not navigation alone, but integrating steering and payload delivery without sacrificing the soft, low-friction behaviour. Internal-channel payload delivery and tip-steering are promising directions, but their co-design within a constrained diameter remains unresolved. Sensing, closed-loop control, and safety validation are essential for clinical translation but are not covered further in this paper; a potential future work.

\section*{\scriptsize ACKNOWLEDGMENT}
\vspace{-0.01\linewidth}
\scriptsize
This work is supported by ERC grant EndoTheranostics, 101118626. Funded by the European Union. Views and opinions expressed are however those of the author(s) only and do not necessarily reflect those of the European Union or the European Research Council Executive Agency. Neither the European Union nor the granting authority can be held responsible for them.

\vspace{-0.01\linewidth}
\scriptsize
\bibliographystyle{IEEEtran}
\bibliography{HSMR}

\end{document}